\documentclass[11pt]{article}
\usepackage{nodalida2025}
\usepackage{times}
\usepackage{url}
\usepackage{latexsym}
\usepackage{graphicx}
\usepackage{amsmath}
\usepackage{hyperref}
\aclfinalcopy 

\title{The Accuracy, Robustness, and Readability of \\ LLM-Generated Sustainability-Related Word Definitions}

\author{Alice Heiman \\
  Stanford University \\
  {\tt aheiman@stanford.edu} \\}

\date{}

\begin{document}
\maketitle
\begin{abstract}
A common language with standardized definitions is crucial for effective climate discussions. However, concerns exist about LLMs misrepresenting climate terms. We compared 300 official IPCC glossary definitions with those generated by GPT-4o-mini, Llama3.1 8B, and Mistral 7B, analyzing adherence, robustness, and readability using SBERT sentence embeddings. The LLMs scored an average adherence of $0.57-0.59 \pm 0.15$, and their definitions proved harder to read than the originals. Model-generated definitions vary mainly among words with multiple or ambiguous definitions, showing the potential to highlight terms that need standardization. The results show how LLMs could support environmental discourse while emphasizing the need to align model outputs with established terminology for clarity and consistency.
\end{abstract}

\section{Introduction}

Large language models (LLMs) have proven effective in a range of tasks, such as analyzing climate-related texts \cite{callaghanMachinelearningbasedEvidenceAttribution2021} and explaining sustainability reports \cite{niCHATREPORTDemocratizingSustainability2023}. However, as citizens and politicians turn to LLMs for information and inspiration, there is concern that these probabilistic models fail to consistently convey the specificity and accuracy required to discuss climate change. For example, agreeing to a standard set of definitions is essential to achieve common ground in the climate debate \cite{ReviewSustainabilityTerms2007}. However, streamlining language around climate is already challenging. For instance, \newcite{ConceptualizingCircularEconomy2017} showed that among 114 different definitions for ``circular economy,'' most failed to convey all nuances of the concept. Thus, this can lead to inconsistencies in research and policy-making.

To address this issue, the Interdisciplinary Panel on Climate Change (IPCC) and the United Nations (UN) maintain the online glossaries IPCC Glossary \cite{IPCC2019, IPCC2019b, IPCC2018}, and UNTERM \cite{unterm_2024}. Although LLMs have access to these repositories during training, they are not constrained to them during inference. Therefore, LLMs could further diversify and confuse these terms. As more people rely on LLMs, it is of special interest to study how LLM-generated explanations adhere to the official definitions, how robust the completions are, and what lessons we should keep in mind when using these models at ever higher levels of climate discourse. Motivated by this, we analyze the adherence, robustness, and readability of word definitions generated by one closed-source and two open-source models compared to official IPCC definitions.

\section{Related Work}

\newcite{phamWordDefinitionsLarge2024} showed that word definitions of English words given by OpenAI LLMs agree well with three popular English dictionaries. However, current LLM performance is mainly dependent on prompt engineering. \newcite{atilLLMStabilityDetailed2024} examined LLM stability and showed that even the same input and parameters can result in variation, which is task-dependent and not normally distributed.

Studies show that sustainability literature can be complex to read \cite{smeuninxMeasuringReadabilitySustainability2020a, barkemeyer_linguistic_2016}. This complexity challenges the accessibility and transparency of sustainability debates and reporting. Studies spanning the sustainability to medical domains use LLMs to simplify these texts and make them interactive \cite{niCHATREPORTDemocratizingSustainability2023, yaoREADMEBridgingMedical2024}.

\section{Methodology}

We present a framework for assessing the adherence and robustness of LLM sustainability word definitions. Specifically, given a term, we let an LLM generate five definitions for each of the five prompt templates (25 completions per term). Then, we use SBERT sentence embeddings to compute the sentence similarity between the official and generated definitions (adherence), as well as the similarity between the generated definitions for a given term and prompt template (robustness). Thus, we define adherence and robustness for each term as follows:

$$\text{adherence} = \frac{1}{n}\sum_{k=1}^n \text{sim}(D, M_k)$$
$$\text{robustness} = \frac{1}{\text{cmb(n)}}\sum_{p=1}^n \sum_{q=k+1}^n  \text{sim}(M_p, M_q)$$

where $D$ is the IPCC glossary definition, $M_k$ is the k’th model definition completion across all prompts, $\text{cmb(n)}$ the number of unique pairwise combinations using $n$ terms, and $\text{sim(A, B)}$ the cosine distance between the SBERT sentence embeddings of the texts A and B. Intuitively, adherence measures how similar model completions are to glossary definitions, while robustness measures the consistency of model completions.



\vspace{5pt}
\noindent\textbf{Dataset collection:} We use Selenium WebBrowser to scrape all terms and definitions from the IPCC glossary website as of December 2024. In total, the glossary contained 911 terms. We limit the terms to those with an overlap in the IPCC 2022 Special Report on Climate Change and Land Annex I Glossary \cite{IPCC2019c}, and get a subset of 300 terms. Finally, we use only the first sentence of each definition and replace all cross-references (such as ``See Pathways'') with the cited term.

\noindent\textbf{Models:} We use three different models in the experiments. We use GPT-4o-mini as our closed source model, and Meta-Llama-3.1-8B-Instruct \cite{llama318B} and Mistral-7B-Instruct-v0.2 \cite{mistral} as our open source models. We use the default parameter settings for all models.

\vspace{5pt}
\noindent\textbf{Prompts:} We prompt ChatGPT with ``Write 4 versions of asking `Define ``[TERM]'' in one sentence.''' resulting in the following list of 5 prompt templates:
\begin{itemize}
    \item Define ``[TERM]'' in one sentence.
    \item How would you define ``[TERM]'' in a single sentence?
    \item Can you describe ``[TERM]'' in just one sentence?
    \item What is your one-sentence definition of ``[TERM]''?
    \item In one sentence, what does ``[TERM]'' mean to you?
\end{itemize}

\vspace{5pt}
\noindent\textbf{Readability analysis:} We use the Python library Readability \cite{Py-Readability-Metrics} to compute the two readability metrics Flesch-Kincaid \cite{kincaid_fishburne_richard_rogers_chissom} and Gunning-Fog \cite{gunning1952technique} for the official definitions and model completions, respectively. Higher Flesh-Kincaid and Gunning-Fog scores indicate more complex material. The metrics require at least 100 words and are not directly applicable to single sentences. Therefore, we use bootstrapping with 1,000 iterations to create longer text samples by sampling 50 random definitions with replacement and assessing the readability of these excerpts.


\section{Experimental Results}

\subsection{Adherence} 
\begin{figure*}
    \centering
    \begin{scriptsize}
    \includegraphics[width=1\linewidth]{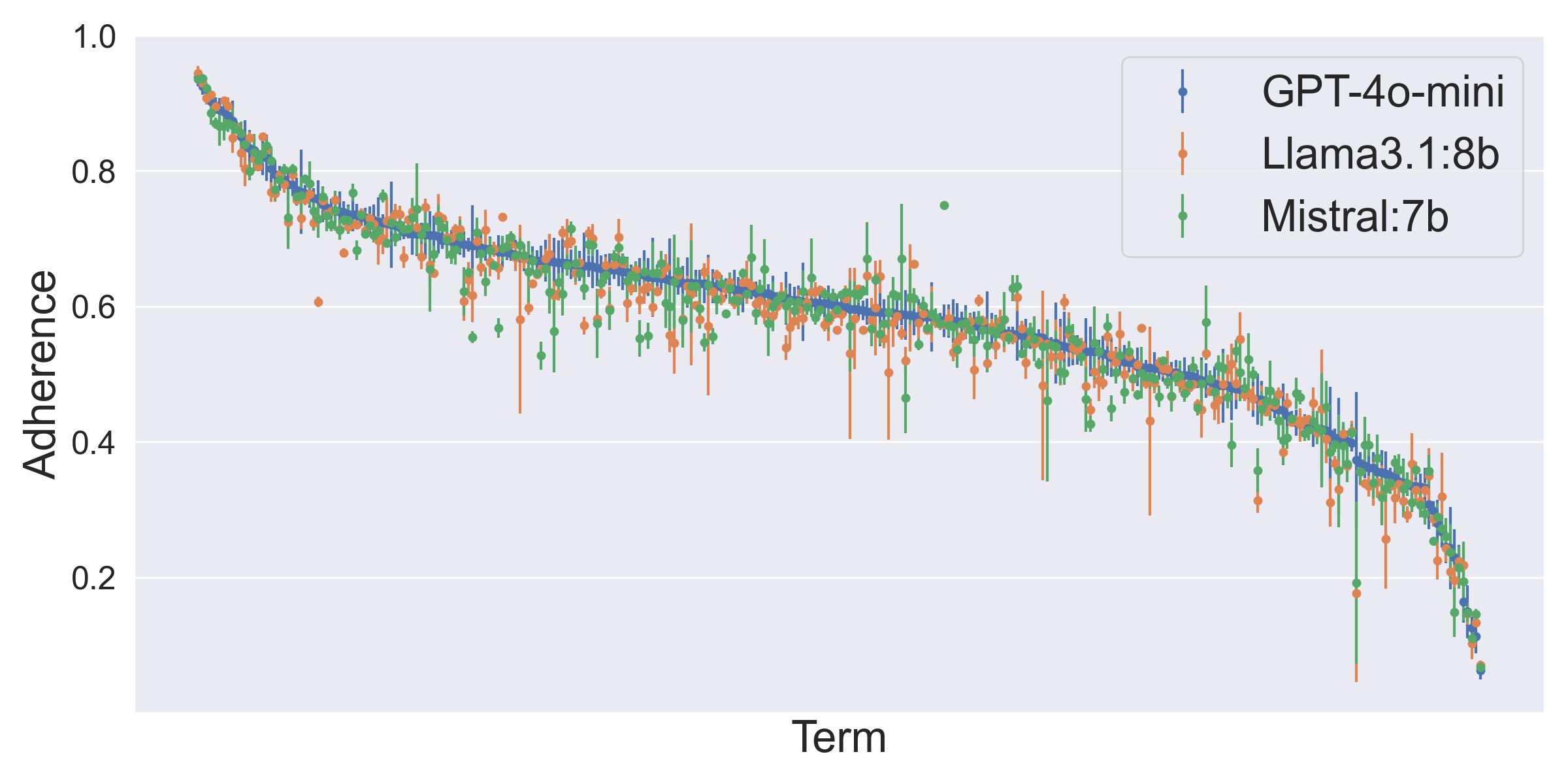}
    \caption{Distribution of SBERT adherence scores between LLM and official IPCC word definitions.}
    \label{fig:term_def_all}
    \end{scriptsize}
\end{figure*}

The average SBERT similarity scores between all terms and their corresponding official IPCC definitions are shown in Figure \ref{fig:term_def_all}. The terms vary greatly, ranging from an adherence score of $0.06$ to $0.94$. Table \ref{table:main_table} shows that all three models received similar results, with average adherence scores of $0.57-0.59 \pm 0.15$. The terms with the highest and lowest adherence scores are shown in Table \ref{table:merged_adherence}. Notably, there is a significant overlap between models, with the term ``East Asian monsoon (EAsiaM)'' scoring highest and ``Demand- and supply-side measures'' scoring lowest.


\begin{table*}[ht]
\centering
\begin{tabular}{|l|c|c|c|c|c|}
\hline
\textbf{Model}         & \textbf{Adherence}               & \textbf{Robustness}               & \textbf{Num Words}       & \textbf{Gunning Fog}      & \textbf{Flesch-Kincaid} \\ \hline
\textbf{GPT-4o-mini}   & $0.59 \pm 0.15$                  & $0.96 \pm 0.02$                  & $34.3 \pm 51.5$          & $22.4 \pm 0.3$           & $19.4 \pm 0.2$          \\ \hline
\textbf{Llama 3.1 8B}  & $0.57 \pm 0.15$                  & $1.00 \pm 0.01$                  & $39.7 \pm 61.4$          & $22.9 \pm 0.3$           & $19.9 \pm 0.2$          \\ \hline
\textbf{Mistral 7B}    & $0.58 \pm 0.15$                  & $1.00 \pm 0.00$                  & $33.6 \pm 69.5$          & $20.8 \pm 0.3$           & $18.1 \pm 0.2$          \\ \hline
\textbf{Definitions}   & -                                & -                                & $30.2 \pm 295.5$         & $19.7 \pm 0.8$           & $16.3 \pm 0.7$          \\ \hline
\end{tabular}
\caption{Adherence, robustness, and readability scores for various LLMs.}
\label{table:main_table}
\end{table*}

\begin{table*}[ht]
\centering
\begin{tabular}{|c|p{6.3cm}|p{6.3cm}|} 
\hline
\small  \textbf{Model}           & \small  \textbf{Highest Adherence Terms}                                      & \small  \textbf{Lowest Adherence Terms}                                      \\ \hline
\small  \textbf{GPT-4o-mini}      & \small 1. East Asian monsoon (EAsiaM)                                & \small 1. Demand- and supply-side measures                            \\ 
                       & \small 2. Eastern boundary upwelling systems (EBUS)                  & \small 2. Poverty                                                     \\
                       & \small 3. Reducing Emissions from Deforestation and Forest Degradation (REDD+) & \small 3. Leakage                                                     \\
                       \hline
\small  \textbf{Llama3.1:8b}        & \small 1. East Asian monsoon (EAsiaM)                                & \small 1. Demand- and supply-side measures                            \\
                       & \small 2. Eastern boundary upwelling systems (EBUS)                  & \small 2. Leakage                                                     \\
                       & \small 3. Deliberative governance                                    & \small 3. Poverty                                                     \\
                       \hline
\small  \textbf{Mistral:7b}      & \small 1. Eastern boundary upwelling systems (EBUS)                  & \small 1. Demand- and supply-side measures                            \\
                       & \small 2. East Asian monsoon (EAsiaM)                                & \small 2. Leakage                                                     \\
                       & \small 3. Reducing Emissions from Deforestation and Forest Degradation (REDD+) & \small 3. Poverty                                                     \\
                       \hline
\end{tabular}
\caption{Terms with the highest and lowest adherence scores between generated and official definitions.}
\label{table:merged_adherence}
\end{table*}

\begin{table}[ht]
\centering
\begin{tabular}{|c|p{4.7cm}|}
\hline
\small  \textbf{Model}           & \small  \textbf{Lowest Robustness Scores}                              \\ \hline
\small \textbf{GPT-4o-mini}      &  \small 1. Projection                                                  \\ 
                           & \small 2. Equity                                                      \\
                           & \small 3. Adaptation pathways                                         \\
                           \hline
\small  \textbf{Llama3.1:8b}      & \small 1. Exposure                                                     \\
                           & \small 2. Glacier                                                      \\
                           & \small 3. Forest                                                       \\
                           \hline
\small \textbf{Mistral:7b}       & \small 1. Sea ice                                                     \\
                           & \small 2. Global mean surface air temperature (GSAT)                  \\
                           & \small 3. Ensemble                                                     \\
                           \hline
\end{tabular}
\caption{Terms with the lowest robustness score between the generated and official definitions.}
\label{table:lowest_robustness}
\end{table}

\subsection{Robustness}
Table \ref{table:main_table} includes the robustness scores across all term completions. The average robustness falls between $0.96-1.00 \pm 0.02$ (min $0.89$, max $1.00$), with no statistical difference between the prompt templates. Some terms produce notable variations, however, in definitions across prompt templates, as listed in Table \ref{table:lowest_robustness}. For instance, GPT-4o-mini's definition of ``Projection'' spanned the psychological (``Projection is a psychological defense mechanism...''), mathematical (``Projection is the process of transferring an image, shape, or data representation...''), and environmental (``Projection" refers to the process of estimating or forecasting future events'') topics. This is to be expected, however, since the prompt did not constrain the model to a particular context. On the other hand, prompting without context gives a hint into potential ambiguities when adapting terms such as ``Equity'', ``Exposure'', and ``Adaptation pathways'' into the climate debate.

\subsection{Readability} 
Table \ref{table:main_table} shows the definitions' average lengths and readability scores. The scores indicate that both IPCC- and model-generated definitions are at the reading level of college graduates. Nevertheless, the IPCC definitions are significantly less complex according to both readability metrics and use fewer words than all model-generated definitions.



\subsection{Ablation Case Studies}

\begin{table*}[ht]
\centering
\begin{tabular}{|l|c|c|c|c|}
\hline
\textbf{Ablation}     & \textbf{Adherence}               & \textbf{Num Words}               & \textbf{Gunning Fog}            & \textbf{Flesch-Kincaid}        \\ \hline
\textbf{LLama+IPCC}          & $0.60 \pm 0.14$                  & $38.1 \pm 45.1$                  & $22.5 \pm 0.3$                 & $19.4 \pm 0.2$                 \\ \hline
\textbf{Llama+Readable}      & $0.55 \pm 0.16$                  & $43.4 \pm 58.6$                  & $19.2 \pm 0.2$                 & $16.4 \pm 0.2$                 \\ \hline
\textbf{Llama+IPCC+Readable} & $0.58 \pm 0.14$                  & $42.6 \pm 47.9$                  & $20.2 \pm 0.2$                 & $17.3 \pm 0.1$                 \\ \hline
\end{tabular}
\caption{Adherence and readability scores for different ablation studies using Llama3.1 8B.}
\label{table:ablation}
\end{table*}

\begin{table*}[ht]
\centering
\begin{tabular}{|l|p{12cm}|}
\hline
\small  \textbf{Ablation}    & \small  \textbf{Radiative Forcing Definition} \\ \hline
\small  \textbf{Official Definition}    & \small  The change in the net, downward minus upward, radiative flux (expressed in W m–2) due to a change in an external driver of climate change, such as a change in the concentration of carbon dioxide (CO2), the concentration of volcanic aerosols or in the output of the Sun. \\ \hline
\small  \textbf{Llama+IPCC}          & \small  Radiative forcing is the change in the balance between the incoming and outgoing radiation at the top of the atmosphere, usually expressed in watts per square meter (W/m²), resulting from a specific perturbation, such as the increase in greenhouse gases. \\ \hline
\small  \textbf{Llama+Readable}      & \small  Radiative forcing is the amount of extra energy that the Earth's atmosphere receives from the sun, usually because of something like pollution or a change in the atmosphere, which can make the planet warmer or cooler. \\ \hline
\small  \textbf{LLama+IPCC-Readable} & \small Radiative forcing is the change in the amount of energy that the Earth's atmosphere receives from the sun, usually caused by human activities or natural changes, which can make the planet warmer or cooler. \\ \hline

\end{tabular}
\caption{Case Study: Ablation study using LLama 3.1 8B to define ``Radiative Forcing'' using three different prompting strategies. ``IPCC'' explicitly asks for a definition in line with the official definition, ``Readable'' for an easily understandable description, and ``IPCC+Readable'' combines the two.} 
\label{table:case_study}
\end{table*}

We perform three additional ablation studies using Llama3.1 8B, using the following prompts:

\begin{itemize}
    \item \textbf{IPCC}: `Define ``[TERM]'' in one sentence. Adhere to the official Intergovernmental Panel on Climate Change (IPCC) glossary without citing it.'
    \item \textbf{Readable}: `Define ``[TERM]'' in one sentence. You must also make the definition understandable by a 10-year old.'
    \item \textbf{IPCC+Readable}: `Define ``[TERM]'' in one sentence. Adhere to the official Intergovernmental Panel on Climate Change (IPCC) glossary without citing it. You must also make the definition understandable by a 10-year old.'
\end{itemize}

\noindent Table \ref{table:ablation} shows the adherence and readability scores using the ablation prompt templates. Notably, the adherence score remains roughly unchanged using the IPCC-specific prompt. Instead, the readability prompt seems to have a greater effect, decreasing the Flesch-Kincaid score from $19.9\pm0.2$ to $16.4\pm0.02$. Although the prompt specified language for a 10-year hold, the Flesh-Kincaid score still corresponds to a college reading level. The relatively high score may partly be explained by the increased sentence length in the LLM's attempt to elaborate and explain parts of the concepts. Table \ref{table:case_study} shows case studies for the term ``Radiative Forcing'' for the official IPCC definition and ablations comparing the definitions generated from different prompts.

\section{Discussion}

The adherence scores suggest that all LLMs generally capture the core semantic meanings of official definitions. Intriguingly, all LLMs achieved similar average adherence scores and had many common outlier terms. This similarity may be due to the models being trained using similar methods and on roughly the same training data. Notably, the adherence score did not significantly improve when we explicitly prompted the model for IPCC definitions. These results imply that providing a climate context may not automatically align language models for a given terminology group. The models do not have perfect recall of definitions; instead, they operate based on probability distributions. Therefore, it is advisable to include the exact definitions in the prompts or LLM systems to ensure they are readily available for reference.


Regarding robustness, the five prompt templates tested did not result in significant variations in generated model definitions. However, there was a notable variability among several terms. As anticipated, the terms with lower robustness scores tend to have multiple meanings, such as ``Projection'', ``Exposure'', and ``Equity''. For instance, ``Equity'' displayed many definitions, reflecting its complex and multi-faceted meanings. This ambiguity aligns with discussions in recent sustainability reports, such as the UN's 2024 Emissions Gap Report, which dedicates an entire section to discuss different equity models \cite{environment_2024}. Thus, the robustness score can help target terms needing further standardization. However, we must also note that robustness is very dependent on the temperature settings of the models. In this paper, we use the default temperature for the models. However, model parameters play a significant role in the consistency and variability of model outputs. These variations could impact how the model presents terms to different users across time. 

In terms of readability, both the IPCC and model definitions scored poorly across both readability metrics. This finding is consistent with previous studies, which suggest that sustainability texts are inaccessible to most readers. Notably, all model completions consistently received more complex readability scores than the already intricate official definitions. This discrepancy may partly be attributed to the longer model responses. Moreover, increasing the readability proved difficult. Although the model used more straightforward terminology, prompting for readability made the model more verbose. Additionally, the readability metrics were not initially designed for single sentences, suggesting that using multiple sentences may yield a more representative assessment.

Future work could explore ways to improve accessibility by using LLMs to simplify language without compromising accuracy and incorporating relevant official glossaries as part of an in-context learning approach. One challenge will be balancing simplicity with accuracy. Adherence scores could offer a helpful framework for evaluating and refining these model outputs since they rely not on exact sentence matching but semantic meaning. Studies across more models and languages would further inform how LLMs represent sustainability.

\section{Conclusion}

This study provides a comprehensive framework for assessing the adherence, robustness, and readability of LLM-generated definitions of sustainability terms compared to official glossaries. While the LLMs capture the semantic meaning of most terms, there is significant variation, particularly for terms with multiple meanings or ambiguous definitions. In addition, IPCC and model definitions show low readability, highlighting the need for further work to simplify sustainability-related language without sacrificing accuracy. Moreover, the case studies show the difficulty in retrieving official definitions even using explicit prompting, indicating the need to include official definitions directly in the prompt. These findings highlight the potential of LLMs to support the environmental conversation but also underscore the importance of carefully aligning model outputs with established terminology to ensure clarity and consistency.

\section*{Acknowledgments}

I want to express my sincere gratitude to Professor Chris Piech for his continued support and encouragement. I also appreciate Maty Bohacek and Anjali Sreenivas's thoughtful comments and invaluable advice.



\bibliographystyle{acl_natbib}
\bibliography{nodalida2025}

\end{document}